\documentclass[10pt,twocolumn,letterpaper]{article}

\usepackage[pagenumbers]{cvpr} 

\usepackage{booktabs}
\usepackage{tabularx}
\usepackage{amssymb}
\usepackage{pifont}
\usepackage[table]{xcolor}
\usepackage{subcaption}
\usepackage{amsmath}
\usepackage{siunitx}
\usepackage{tikz}
\usepackage{siunitx}
\usepackage{nicematrix}
\usepackage{pifont}        
\newcommand{\cmark}{\ding{51}}
\newcommand{\xmark}{\ding{55}}
\newcommand{\best}[1]{{\bfseries #1}}

\usepackage[accsupp]{axessibility}  

\usepackage{booktabs}
\usepackage{tabularx}
\usepackage{array}
\newcolumntype{Y}{>{\centering\arraybackslash}X}
\usepackage{booktabs,tabularx,array,pifont}
\usepackage[table]{xcolor} 
\newcolumntype{Y}{>{\centering\arraybackslash}X}

\definecolor{cvprblue}{rgb}{0.21,0.49,0.74}
\usepackage[pagebackref,breaklinks,colorlinks,allcolors=cvprblue]{hyperref}

\usepackage[capitalize]{cleveref}
\crefname{section}{Sec.}{Secs.}
\Crefname{section}{Section}{Sections}
\Crefname{table}{Table}{Tables}
\newcommand{\Tref}[1]{Table~\ref{#1}}

\newcommand{\Fref}[1]{Fig.~\ref{#1}}

\def\paperID{5548} 
\def\confName{CVPR}
\def\confYear{2026}

\title{Stay in your Lane: \\
Role Specific Queries with Overlap Suppression Loss for Dense Video Captioning
}

\author{
    Seung Hyup Baek$^{1*}$\quad Jimin Lee$^{2*}$\quad Hyeongkeun Lee$^3$\quad Jae Won Cho$^{1\dagger}$ \\[2mm]
    $^1$Konkuk University \quad $^2$Sejong University \quad $^3$KAIST \\
    {\tt\small \{edwardback, chojw\}@konkuk.ac.kr, jjujimin@sju.ac.kr, lhk528@kaist.ac.kr}
}

\begin{document}
\maketitle
\def\thefootnote{*}\footnotetext{Equal contribution}
\def\thefootnote{$\dagger$}\footnotetext{Corresponding author}
\def\thefootnote{\arabic{footnote}}
\begin{abstract}

Dense Video Captioning (DVC) is a challenging multimodal task that involves temporally localizing multiple events within a video and describing them with natural language. While query-based frameworks enable the simultaneous, end-to-end processing of localization and captioning, their reliance on shared queries often leads to significant multi-task interference between the two tasks, as well as temporal redundancy in localization.
In this paper, we propose utilizing role-specific queries that separate localization and captioning into independent components, allowing each to exclusively learn its role.
We then employ contrastive alignment to enforce semantic consistency between the corresponding outputs, ensuring coherent behavior across the separated queries.
Furthermore, we design a novel suppression mechanism in which mutual temporal overlaps across queries are penalized to tackle temporal redundancy, supervising the model to learn distinct, non-overlapping event regions for more precise localization.
Additionally, we introduce a lightweight module that captures core event concepts to further enhance semantic richness in captions through concept-level representations.
We demonstrate the effectiveness of our method through extensive experiments on major DVC benchmarks YouCook2 and ActivityNet Captions. Code is available \href{https://github.com/MMAI-Konkuk/ROS-DVC}{here}.

\end{abstract}

\begin{figure}[t]
    \centering
        \includegraphics[width=1.0\linewidth]{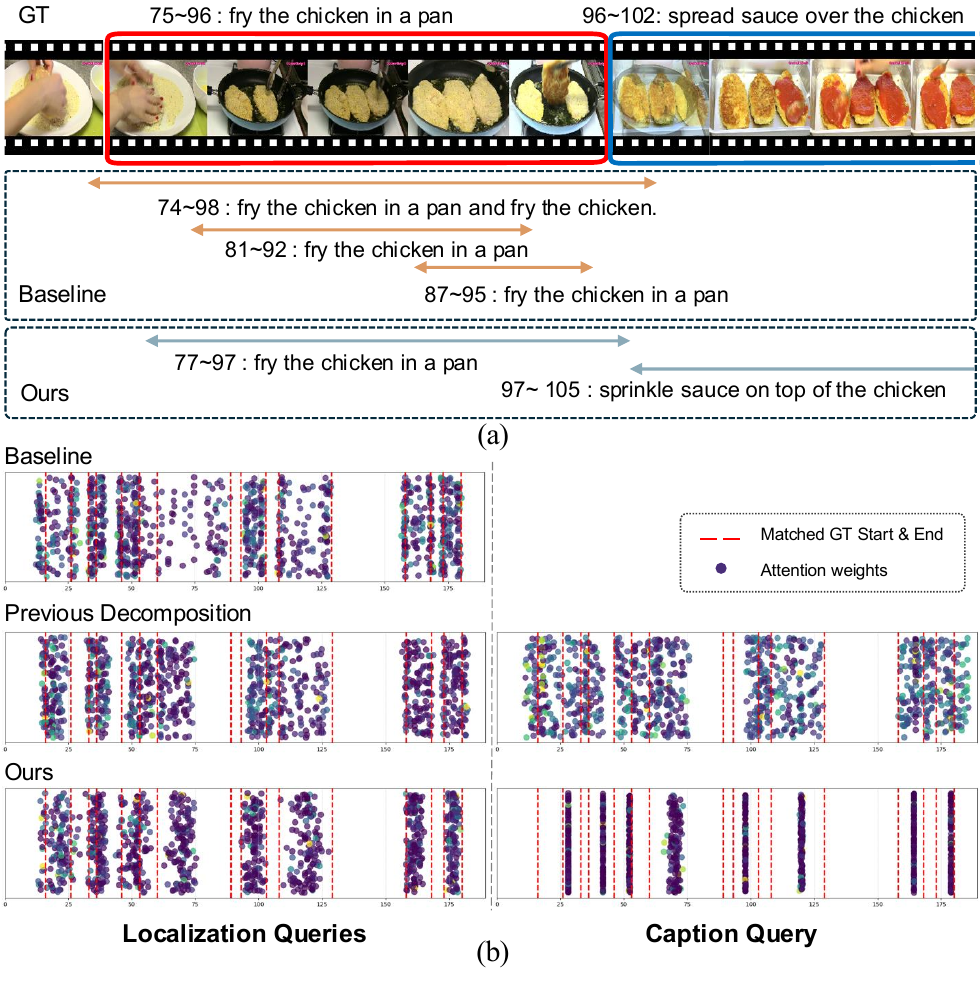}
    \caption{Comparison of our method with previous methods.
(a) Captioning results. The baseline model captures overlapping events, causing redundant captions. In contrast, our model generates distinct non overlapping regions for a set caption.
(b) Decoder query sampled attention weights. The baseline's single query attends indiscriminately. Previous ``decomposition'' queries show similar attention distributions. In contrast, our method employs two specialized queries: localization queries attend broadly for boundaries and a caption queries attend densely on key frames.}
    \label{fig:fig1}
\end{figure}

\newcommand{\modelname}{[Our Model Name]} 

\section{Introduction}
\label{sec:introduction}

With the explosive growth of online video content, automatic video understanding~\cite{zhou2023procedure_vidund1,wang2023internvid_vidund2,song2024moviechat_vidund3, wang2023selective_vidund4,mangalam2023egoschema_vidund5,li2024videomamba_vidund6,li2024mvbench_vidund7,woo2024let_cite1,jang2023self_cite2,jang2022signing_cite3} has emerged as a central challenge in computer vision and natural language processing. Through these efforts, video captioning~\cite{wang2018reconstruction_vidcap1,qi2019sports_vidcap2,pei2019memory_vidcap3,gao2017video_vidcap4,chen2017video_vidcap5} has drawn significant attention for its ability to translate video content into natural language. 

Most conventional video captioning approaches focus on short, trimmed clips. However, these methods struggle to generalize to real-world videos, which are often long, untrimmed, and contain multiple events.
To address this limitation, Dense Video Captioning (DVC)~\cite{Krishna2017dvc_dvc1} was introduced as a task. DVC requires not only the precise temporal localization of all salient events but also the generation of coherent textual descriptions for each localized segment.


Traditional DVC methods have adopted a two-stage ``localize-then-describe'' approach~\cite{Krishna2017dvc_dvc1, li2018jointly_task_inference_1,wang2018bidirectional_two_stage2,iashin2020multi_two_stage1}, which often suffered from not being end-to-end trainable and led to suboptimal performance from the lack of interaction between the two seminal tasks. 
Subsequently, PDVC~\cite{Wang2021pdvc_dvc2} first leveraged the DETR~\cite{2020detr} network 
and introduced a unified query-based framework where a fixed set of learnable queries jointly predict event segments and generate captions through parallel decoding. This design enabled end-to-end joint optimization of both tasks and significantly simplified the DVC pipeline. Building upon this paradigm, subsequent research~\cite{kim2024cm2_dvc3,xie2025exploringmccl_dvc4,Wu_2025_CVPRe2dvc_dvc5,liu-etal-2025-taskddvc_dvc6} has continued to improve performance within this DETR-based architecture.

However, 
utilizing a shared set of learnable queries for both localization and captioning simultaneously hinders
the model’s ability to capture the distinct characteristics of each task. 
Training a single query to handle both objectives has been shown to be a possible cause of task interference in general multi-task learning contexts, where the shared representation space leads to conflicting optimization directions~\cite{li2018jointly_task_inference_1,chen2023task_task_inference_2,yan2024multi_task_inference_3,ding2023mitigating_task_inference_4,qin2025towards_task_inference_5}.
We observe a similar issue in query-based DVC models. 
As shown at the top of \Fref{fig:fig1}(b), the baseline PDVC~\cite{Wang2021pdvc_dvc2} exhibits ambiguous query behaviors, with attention weights that fail to focus either on precise event boundaries or on semantically coherent regions for captioning. 
Although DDVC~\cite{liu-etal-2025-taskddvc_dvc6} introduces query decomposition, we observe that, in the center of~\Fref{fig:fig1}(b), the attention weights of its localization and caption queries appear highly similar, as they do not create separate queries but instead utilize an MLP, suggesting that proper task separation has not been effectively achieved. In addition, 
as shown in \Fref{fig:fig1}(a), we observe that queries tend to repeatedly capture similar temporal segments and scenes, generating redundant captions. This behavior degrades both localization and captioning performance.
To overcome these challenges, we argue that a more direct supervision is necessary to properly train the query learning process. To achieve this, we assert that queries must: 1) resolve task interference through
clear task separation
and 2) learn to avoid overlapping temporal regions. However, 
task separation can lead to misalignment between the two query types. Therefore, 
the queries must not only be separated and trained to be non-overlapping, but also be trained to be semantically coherent across the separated tasks through a cross-task supervision loss. 

In this paper, we propose \textbf{R}ole Specific Query with \textbf{O}verlap \textbf{S}uppression \textbf{D}ense \textbf{V}ideo \textbf{C}aptioning (\textbf{ROS-DVC}), which tackles the aforementioned issues through three core techniques. We introduce role specific queries, where we separate localization and caption queries into independent query sets and initialize them each from separate
learnable embedding spaces. In our role specific queries, as shown at the bottom of~\Fref{fig:fig1}(b), the
localization queries broadly attend to temporal context to predict boundaries, while the caption queries focus intensively around a key region to describe its semantics in detail. To ensure semantic alignment between these decoupled representations, we introduce the Cross-Task Contrastive Alignment (CTCA) Loss. 
This loss ensures the coupling between the two tasks by aligning localized events and captions together. 
To mitigate the query overlap problem, we introduce a novel Overlap Suppression Loss that explicitly penalizes query pairs that overlap excessively based on pairwise IoU between the queries. 
This encourages the localization queries to focus on distinct regions and ensure precise localization.
To further enhance the semantic expressiveness of captions, we introduce a Concept Guider, a lightweight MLP-based concept head that captures the underlying concepts~\cite{lu2024set_concept} of events. This simple yet effective module enriches the caption queries with concept-level representations, enabling the model to generate captions that better reflect the semantic understanding of each event.

Our main contributions can be summarized as follows:
\begin{itemize}
    \item We resolve task interference in DVC by utilizing independent, role specific queries trained with our CTCA Loss. 
    \item 
    We design an Overlap Suppression Loss that discourages unnecessary query overlaps, enabling the model to learn more accurate temporal localizations.
    \item We introduce a Concept Guider that enhances the expressiveness and semantic accuracy of generated captions 
    without utilizing an external memory bank.
    \item We achieve competitive performance on DVC benchmarks YouCook2~\cite{youcook2} and ActivityNet Captions~\cite{Krishna2017dvc_dvc1}.
\end{itemize}

\begin{figure*}[t]
    \centering
    \includegraphics[width=1.0\linewidth]{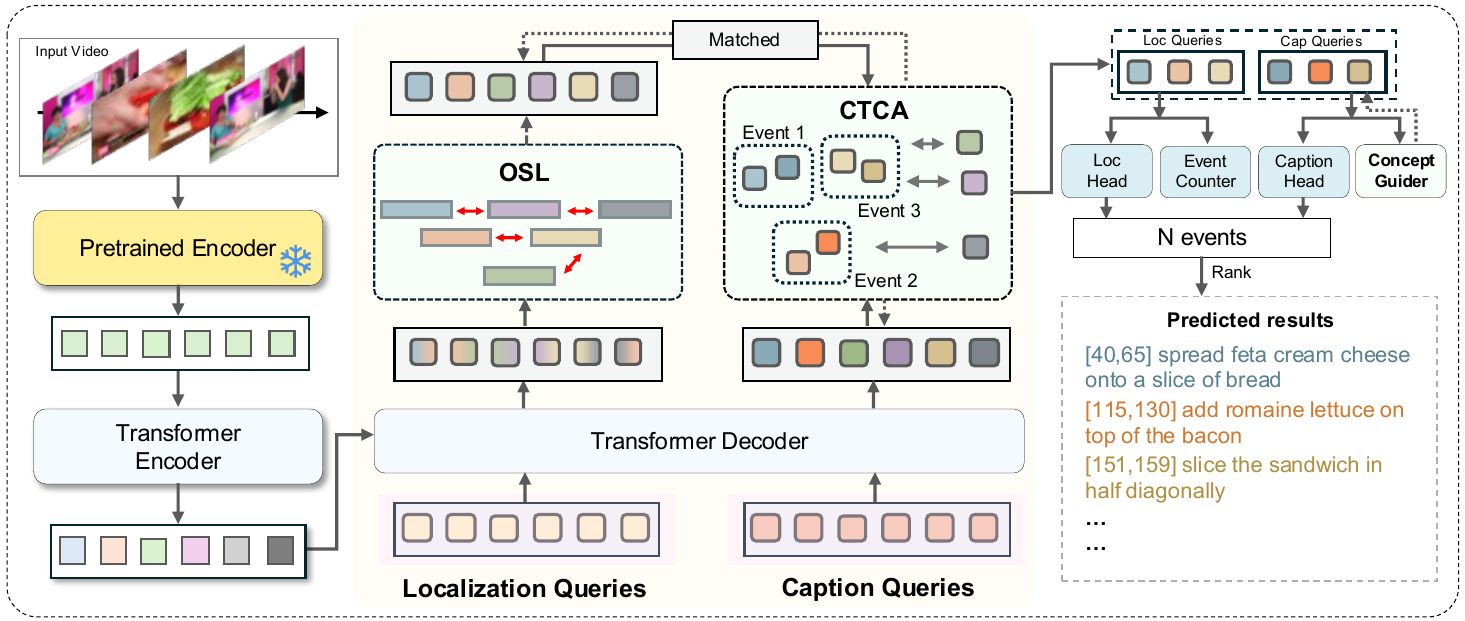}
    \caption{\textbf{An overview of our proposed ROS-DVC framework.} The input video is first fed into the pretrained encoder, and a transformer encoder processes it to generate frame-level features. In the decoding stage, two types of queries are independently initialized and retrieve their role-specific information from the frame-level features. The output localization queries are trained with the Overlap Suppression Loss to minimize mutual overlap and are matched with ground truths via the Hungarian algorithm. Subsequently, the CTCA loss is employed to semantically align the caption queries with their corresponding localization queries. Finally, these processed queries are fed into respective heads to obtain the predictions for event-number, localize timestamps, event captions, and event-level concepts.}
    \label{fig:framework}
\end{figure*}

\section{Related Works}
\label{sec:related}


\noindent\textbf{Dense Video Captioning} (DVC)~\cite{Krishna2017dvc_dvc1} is a multimodal task comprising two sub-tasks: {event localization} and {captioning}. Early works adopt numerous approaches to tackle this problem~\cite{wang2018bidirectional_two_stage2,wang2020event_two_stage3,ryu2021semantic,rahman2019watch_e2e3,iashin2020multi_two_stage1,chang2022event,aafaq2022dense}. 
The pioneering work of DVC~\cite{Krishna2017dvc_dvc1} has utilized a two-stage ``localize-then-describe'' approach~\cite{wang2018bidirectional_two_stage2,wang2020event_two_stage3,iashin2020multi_two_stage1}, where it first identifies the temporal location of an event and then feed its boundaries into a separate captioning head to generate a description. 
However, Wang~\etal~\cite{Wang2021pdvc_dvc2} have criticized this line of research for its heavy reliance on anchor-based designs to improve localization and the inherent lack of mutual interaction between the two sub-tasks.
To address these limitations, later studies began to explore end-to-end frameworks~\cite{zhou2024streaming_e2e1,yang2023vid2seq_e2e2, rahman2019watch_e2e3, mun2019streamlined_e2e4, jeon2025sali4vid_cite4} that jointly optimize event localization and captioning within a unified architecture, allowing the two sub-tasks to interact more closely during training and inference.
Due to the similarity of object detection's task structure and that of DVC, both of which involve the two core components of localization and semantic understanding, 
~\cite{Wang2021pdvc_dvc2} first adopted the DETR architecture to improve efficiency by processing localization and captioning concurrently through a parallel decoding framework.  
Subsequent methods~\cite{kim2024cm2_dvc3, xie2025exploringmccl_dvc4, Wu_2025_CVPRe2dvc_dvc5, liu-etal-2025-taskddvc_dvc6} have improved upon the DETR-based end-to-end pipeline. 

Among recent methods, a line of works has sought to improve the captioning ability of DVC models through a Retrieval-Augmented Generation method, where an external memory bank is used to better align textual and visual modalities for more diverse captions. CM$^2$~\cite{kim2024cm2_dvc3} first proposed the use of an external memory bank, 
and 
MCCL~\cite{xie2025exploringmccl_dvc4}, pointed out the need for synergy between the captioning and localization heads and proposed a cyclic co-learning strategy. 
Another line of work focused on improving the representation of queries. E$^2$DVC~\cite{Wu_2025_CVPRe2dvc_dvc5} focused on query initialization to mitigate data bias and ensure a more uniform detection of events, while DDVC~\cite{liu-etal-2025-taskddvc_dvc6} employed query decomposition through an MLP to make the captioning and localization features more task-specific. 

Most recently, Chen et al.~\cite{chen2025decouplingmm} further explore query decomposition by introducing an additional learnable encoder prior to query initialization to enrich query representations. 
By enhancing query representations through encoder processing, their approach aims to provide more informative features for downstream localization and captioning. 
In contrast, our approach does not rely on additional encoders and instead emphasizes explicit query separation at initialization, focusing on improving the learning of task-specific queries themselves.
We place our work in this line of research, where we tackle DVC through a general query-based framework without the use of an external memory bank.
Our approach enhances the learning of queries themselves by introducing independent task-specific queries and novel loss functions that enable adaptive and disentangled optimization of localization and captioning.

\noindent\textbf{Overlap Suppression.} 
Localization tasks often suffer from overlapping predictions, which are typically mitigated through Non-Maximum Suppression (NMS).
However, more complex tasks like crowd-scene detection adopt various additional regularization methods. 
Wang \etal ~\cite{wang2018repulsion_loss_related4} proposed the repulsion loss to push the predicted bounding box farther away from other predicted bounding boxes and other ground truths. Additionally, AutoPedestrian~\cite{tang2021autopedestrian_loss_related5} introduced an automatic scheme that jointly searches for optimal data augmentation policies and loss function parameters, aiming to improve pedestrian detection under crowded conditions.
Gao \etal.~\cite{gao2023selecting_loss_related6} identified the problem of excessive false-positive predictions in the DETR architecture and proposed a learnable sample selection strategy applied to the matching cost and loss function.
While DVC adopts the DETR architecture and does not use NMS, it still suffers from excessive overlapping predictions.
Drawing inspiration from these approaches, we redesign the loss for our task, while employing a DETR-based architecture despite the task differences.
To address temporal overlap, we propose our Overlap Suppression Loss that leverages the IoU between proposed events and their ground truths, penalizing overlapping proposals and promoting more precise event boundaries.

\label{sec:methodology}
\section{Methodology}

Our method aims to enhance both event localization and captioning by supervising the learning process to capture the distinct characteristics of both detection and captioning. To this end, we introduce three key components: Role Specific Query~\ref{sec:methodrole}, Overlap Suppression Loss~\ref{sec:methoddensity}, and Concept Guider~\ref{sec:methodconcept}, which address task-aware query optimization, suppression of redundant event proposals, and concept-level caption reasoning respectively.
Built upon a parallel encoder–decoder architecture, our framework, the \textbf{ROS-DVC}
method, employs separate localization and caption queries within a shared decoder to jointly refine temporal boundaries and event descriptions. 

\subsection{Overall Framework}
\label{sec:overallframework}


Overall, the DVC model first encodes an input video using a pre-trained visual encoder for frame features and a transformer encoder for temporal context.
In the decoding stage, a set of learnable queries interacts with the encoded frame features to generate $N$ event-caption pairs, ${\{(t_s^n, t_e^n, \text{Cap}^n)\}}_{n=1}^N$, where $N$ denotes the number of events detected in the video.
During training, the Hungarian algorithm~\cite{kuhn1955hungarian} is employed to find optimal one-to-one matchings between $N$ generated pairs and the ground-truth events.
These queries function as learnable representations and act as event proposal candidates.
Each query is associated with a reference point, an initial guess for the event's features and location.
Both the query features and reference points are then iteratively refined via cross-attention across decoding layers, progressively improving the precision of event locations and descriptions.

\subsection{Role Specific Queries}
\label{sec:methodrole}
\subsubsection{Role Specific Query Initialization}
Unlike previous works that use a single set of queries for both sub-tasks, 
we intend localization queries to capture broad temporal context for precise event boundary estimation, whereas caption queries focus on fine-grained details for accurate visual description.
Hence, we explicitly separate the query space into two sets, each specialized for \textit{localization} and \textit{captioning}, respectively.
\begin{equation}
\{ q_{\text{loc}}^j \}_{j=1}^{K}, \quad
\{ q_{\text{cap}}^j \}_{j=1}^{K}.
\end{equation}
Here, $\{ q_{\text{loc}}^j \}$ denotes $K$ localization queries, and $\{ q_{\text{cap}}^j \}$ denotes $K$ caption queries. Unlike~\cite{liu-etal-2025-taskddvc_dvc6}, which generates caption queries from localization queries through an MLP, creating dependencies and limiting the ability to learn more diverse attention distributions,
our approach initializes each query set from its own separate embedding space.
This allows each query to be independently optimized for its own role specific objective while reducing dependencies.

Although the queries are initialized from separate embedding spaces, they maintain complementarity by referencing the same visual location, as they are injected together into the cross-attention layers.
This design ensures that the queries can learn rich semantic representations independently while still being grounded in the relevant visual context defined by the `localization' query. This allows the two tasks to learn at different temporal abstraction levels without conflict and converge to their respective roles.

\subsubsection{Cross-Task Contrastive Alignment}
Following our Role Specific Query Initialization, the model employs localization queries to precisely estimate event boundaries, and $N$ events are selected via the Hungarian algorithm. Next, each of these localization queries must be paired with a semantically matched caption query. However, because caption queries are specialized for capturing semantic meaning and are also initialized from a different embedding space,
their semantic consistency with its corresponding localization queries might not be guaranteed.

To address this issue, we introduce a Cross-Task Contrastive Alignment (CTCA) Loss that bridges the gap between the two obtained queries. 
The output query features are denoted as $ \tilde{ q}_{\text{loc}}$, $ \tilde{ q}_{\text{cap}}$.
In CTCA, the Hungarian algorithm first identifies a set of indices $\mathcal{M}$ corresponding to the ground-truth events. For each $j \in \mathcal{M}$, the $j$-th caption query ($\tilde{q}_{\text{cap}}^j$) and the $j$-th localization query ($\tilde{q}_{\text{loc}}^j$) are treated as positive pairs.
The $j$-th caption query is then contrasted with all other localization queries ($\tilde{q}_{\text{loc}}^{j'}$, where $j' \neq j$) as negatives. This formulation encourages localization queries to be pulled closer to their corresponding caption representations (reinforcing event semantics) while being pushed away from unrelated ones. This, in turn, enables the localization query to effectively learn semantic cues from the caption query.
The CTCA Loss is defined as follows:
\begin{equation}
\mathcal{L}_{\text{CTCA}}
= - \sum_{j \in \mathcal{M}}
\log
\frac{\exp(\text{sim}(\tilde q_{\text{cap}}^j, \tilde q_{\text{loc}}^j)/\tau)}
{\sum_{j'} \exp(\text{sim}(\tilde q_{\text{cap}}^j, \tilde q_{\text{loc}}^{j'})/\tau)},
\end{equation}
Where $\text{sim}(\cdot)$ denotes cosine similarity and $\tau$ is the temperature parameter.
Through this asymmetric contrastive objective, the localization queries acquire semantic awareness from their corresponding captions.
Consequently, the two query sets learn in distinct yet cooperative embedding spaces, one capturing event semantics and the other focusing on temporal structure, leading to more coherent and event-consistent dense video captions.


\subsection{Overlap Suppression Loss (OSL)}
\label{sec:methoddensity}
To detect more precise event boundaries, we introduce the Overlap Suppression Loss (OSL), a loss that induces suppression of the distribution of predicted event boundaries. 
This loss is designed to solve two key challenges: first, to prevent the model from producing excessively overlapping and redundant event predictions, and second, to discourage it from collapsing all boundaries into 
a non-informative solution. 
This helps reduce excessive overlap predictions

The foundation of our loss is the overlapping measure $P_o$, which we define as the temporal 
IoU~\cite{rezatofighi2019generalized_iou} between any two distinct predicted boundaries, $B_i$ and $B_j$:
\begin{equation}
P_o(i,j) = \operatorname{IoU}(B_i, B_j), \quad i \neq j.
\end{equation}
Minimizing $P_o$ alone gives an 
aggressive penalty on all overlaps and could inadvertently suppress multiple correct predictions that are close to one another, thereby limiting the model's ability to identify genuine events. Therefore, relying on it alone can be insufficient.

\begin{figure}[t]
    \centering
    \includegraphics[width=1.0\linewidth]{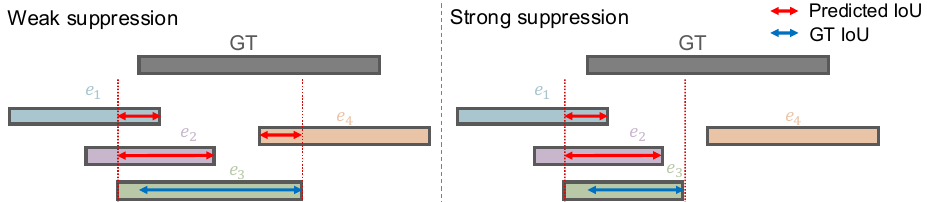}
    \caption{\textbf{Mechanism of the Overlap Suppression Loss.} This loss discourages overlap among queries. The suppression strength is inversely modulated by the query's IoU with the ground truth (GT). Queries with high GT IoU receive weak suppression, while queries with low GT IoU receive strong suppression.}
    \label{fig:OSL}
\end{figure}

To address this, we incorporate a ground-truth–aware weighting factor, $\alpha$, which adaptively modulates the overlap penalty based on the predicted alignment with the ground truth. 
We first quantify this alignment by calculating the ground-truth overlap $P_g$, defined as the IoU between a predicted boundary $B_i$ and the ground-truth boundary $G_j$:
\begin{equation}
P_g(i,j) = \operatorname{IoU}(B_i, G_j).
\label{eq:beta}
\end{equation}
We then define $\alpha$ as the combination of the alignment score $P_g$ with its complement $(1-P_g)$, where the balance between the two is controlled by a hyperparameter $\gamma$. Thus, $\alpha$ is defined as:
\begin{equation}
\alpha = \gamma \cdot P_g + (1 - \gamma) \cdot (1 - P_g).
\label{eq:alpha}
\end{equation}
We let $\gamma \le 0.5$. Given this constraint, a higher $P_g$ leads to a smaller $\alpha$, thereby decreasing the penalty when the prediction is well aligned with the ground truth. Conversely, when $P_g$ decreases, $\alpha$ becomes larger, resulting in a stronger penalty, effectively discouraging the model from predicting overlapping predictions. The final OSL formulation is:
\begin{equation}
L_{\text{OSL}} = -\alpha \cdot \log(\beta - P_o).
\label{eq:density_loss}
\end{equation}
With $\beta$ being a hyperparameter that defines the maximum overlap. 
$\alpha$ allows the model to distinguish between relevant and redundant overlaps, as illustrated in~\Fref{fig:OSL}. 
This mechanism achieves a critical balance between reducing redundancy and ensuring accurate localization, leading to a more precise set of event predictions.

\subsection{Concept Guider}
\label{sec:methodconcept}
Previous research has shown that utilizing concepts can enhance semantic richness of generated captions in DVC~\cite{lu2024set_concept, yang2023concept_concept_2, lu2024set_concept_3, xie2025exploringmccl_dvc4}.
Hence, we introduce the Concept Guider, a lightweight MLP auxiliary head designed to output an event-level multi-hot vector $\hat y_c$, that assists the expressive power of the caption queries.

\begin{equation}
\hat y_c = sigmoid(MLP(\tilde{ q}_{\text{cap}}))
\end{equation}
This head takes the final embedding of the caption query $ \tilde{ q}_{\text{cap}}$ as input and is trained with an auxiliary objective that encourages the query representation to internalize the concepts associated with each event.
Specifically, we extract the top $N_c$ nouns and verbs from training captions, and define them as concepts. Then, for each event, a multi-hot concept label $Y^c \in \{0,1\}^{N_c}$ is constructed, where each element is set to 1 if the relevant concept appears in the ground-truth captions, and 0 otherwise. 
Through this auxiliary task, the caption query is guided to embed high-level semantics and implicit intentions within its representation, moving beyond simple visual features.
Consequently,
the Concept Guider enriches the query embedding, helping the main caption generator produce more specific and contextually aware sentences. This auxiliary head is trained using the Cross-Entropy Loss and is unused during inference.


\subsection{Task Heads}
\label{sec:taskheads}

Our method employs a role specific query based framework, in which the decoder’s final representations are distributed across four specialized heads including the Concept Guider in section ~\ref{sec:methodconcept}. Each head is responsible for a distinct component of dense video captioning. 

\noindent\textbf{Localization Head}
predicts temporal boundaries and assigns confidence scores
for each localization query $\tilde{ q}_{\text{loc}}^i$.
The output is a set of tuples $\{t_i^s, t_i^e, c_i\}_{i=1}^{N_q}$, where $t_i^s$ and $t_i^e$ represent start and end times, and $c_i$ is localization confidence.

\noindent\textbf{Captioning Head}
is based on an LSTM~\cite{hochreiter1997long_lstm}.
At each time step $t$, the LSTM receives context features $z_{i,t}$, caption query $\tilde{ q}_{\text{cap}}^i$, and sequence of previously generated words $\{w_{i,j}\}_{j=1}^{t-1}$.
Then, it outputs the current word $w_{i,t}$.
Through repetition, the complete description is generated as
$S_i = \{w_{i,1}, w_{i,2}, \dots, w_{i,T}\}$,
where $T$ is the sentence length.

\noindent\textbf{Event Counter}
counts the number of events in the video.
The localization query $\tilde{ q}_{\text{loc}}^i$ is transformed using an MLP to
$r^{len}$, which represents the predicted distribution over possible event counts.
During inference, the estimated number of events is obtained as
$N = \text{argmax}(r^{len})$. Subsequently, the $N$ predicted sets $\{(t_n^s, t_n^e, S_n)\}_{n=1}^{N}$
are selected via 
the Hungarian algorithm~\cite{kuhn1955hungarian} using a matching cost defined as $C = \mathcal{L}_{\text{cls}} + \alpha_{\text{giou}}\mathcal{L}_{\text{giou}} + \alpha_{\text{cap}}\mathcal{L}_{\text{cap}}$.
The focal loss $L_{\text{cls}}$ 
measures the difference between the
predicted classification scores and the ground truth labels,
while the generalized IoU loss $L_{\text{giou}}$ measures the discrepancy between the predicted boundary and ground truth temporal segments. The caption loss $L_{\text cap}$ evaluates the quality of the generated captions through cross-entropy.
This extended matching strategy allows caption queries to directly influence the pairing stage, ensuring that each localization \& caption pair is both temporally accurate and semantically consistent.

\subsection{Loss Function}
\label{sec:lossfunction}

For the standard DVC losses, we use $\mathcal{L}_{\text{giou}}$, $\mathcal{L}_{\text{cls}}$, $\mathcal{L}_{\text{cap}}$, and $\mathcal{L}_{\text{ec}}$, where $\mathcal{L}_{\text{ec}}$ is the cross-entropy loss between the predicted event count distribution and the ground-truth count. In addition, we add $\mathcal{L}_{\text{CTCA}}$, $\mathcal{L}_{\text{OSL}}$, and $\mathcal{L}_{\text{CG}}$, with $\mathcal{L}_{\text{CG}}$ defined as the cross-entropy loss between the event-level concept labels and the predicted concept distribution. 

Therefore, our final loss is formulated as follows:
\begin{equation}
\begin{split}
\mathcal{L}_{\text{total}} = & \lambda_{\text{giou}}\mathcal{L}_{\text{giou}} + \lambda_{\text{cls}}\mathcal{L}_{\text{cls}} 
+ \lambda_{\text{cap}}\mathcal{L}_{\text{cap}} + \lambda_{\text{ec}}\mathcal{L}_{\text{ec}} \\
&\hspace{-1.2cm} + \lambda_{\text{CTCA}}\mathcal{L}_{\text{CTCA}} 
+ \lambda_{\text{OSL}}\mathcal{L}_{\text{OSL}} 
+ \lambda_{\text{CG}}\mathcal{L}_{\text{CG}}
\end{split}
\label{eq:total_loss}
\end{equation}
where each $\lambda$ is a hyperparameter that balances the contribution of its respective loss term.

\section{Experiments}
\label{sec:experiments}

\subsection{Experimental settings}
\noindent\textbf{Datasets.} We evaluate our proposed method on two widely used benchmark datasets for Dense Video Captioning (DVC): YouCook2 \cite{youcook2} and ActivityNet Captions \cite{Krishna2017dvc_dvc1}. YouCook2 consists of 2k untrimmed videos of cooking procedures from 89 distinct recipes. The videos have an average length of about 320 seconds and contain an average of 7.7 annotated segments, each with a corresponding sentence description. For our experiments, we adhere to the official split of 1,333 videos for training, 457 for validation, and 210 for testing. 
ActivityNet Captions contains 20k untrimmed videos covering a wide range of human activities, with an average duration of approximately 120 seconds. Each video is annotated with an average of 3.65 temporally localized sentences, resulting in a total of over 100k annotations. We follow the official data split, using 10,009, 4,925, and 5,044 videos for training, validation, and testing, respectively. 
We only utilize videos that remain accessible on YouTube, which accounts for approximately 7\% fewer videos than the original dataset splits. 

\noindent\textbf{Implementation Details. }
For both datasets, video frames are sampled at 1 FPS. The resulting frame sequences are then subsampled or padded to a fixed length $F$, which is set to 200 for YouCook2 and 100 for ActivityNet. We employ CLIP ViT-L/14~\cite{dosovitskiy2020image_vit,radford2021learning_clip} to extract visual features from the video frames, following the standard protocol used in previous works~\cite{kim2024cm2_dvc3}.
Following prior DVC frameworks, we adopt a 2-layer deformable transformer decoder with multi-scale deformable attention across four feature levels. The number of localization and captioning queries are set to 50 each for YouCook2 and 10 each for ActivityNet Captions.
For our proposed modules, we set the hyperparameters $\gamma = 0.25$, $\beta = 1.0$, and $N_C=30$.

\noindent\textbf{Evaluation metrics.}
We evaluate our method on two sub-tasks within DVC: captioning performance and event localization. Following standard practice, all metrics are averaged over IoU thresholds of $\{0.3, 0.5, 0.7, 0.9\}$ to ensure a robust evaluation across various degrees of temporal overlap. For captioning performance, we use the official ActivityNet Challenge evaluation tool to compute scores for 
CIDEr~\cite{vedantam2015cider}, METEOR~\cite{banerjee2005meteor}, BLEU-4~\cite{papineni2002bleu}, and SODA$\_$c~\cite{fujita2020soda}.
For event localization, we evaluate performance using average precision, average recall, and F1-score, which is the harmonic mean of precision and recall.

\begin{table}[t]
\centering
\small
\setlength{\tabcolsep}{0pt} \renewcommand{\arraystretch}{1.1}
\caption{\textbf{YouCook2 Caption metrics.} \textbf{Bold} represents best and \underline{underline} represents second best performance. We do not directly compare to DDVC for fair comparison as they utilize GPT2 as their captioning head.}
\label{tab:yc2_cap}
\begin{tabular*}{\columnwidth}{
    @{\hspace{4pt}}l@{\hspace{4pt}} 
    !{\vrule width 0.5pt} 
    @{\hspace{4pt}}c@{\hspace{4pt}} 
    !{\vrule width 0.5pt} 
    @{\extracolsep{\fill}}
    c 
    c 
    c 
    @{\hspace{4pt}}c@{\hspace{4pt}} %
} 
\toprule
Method & \#PT
& CIDEr\,\(\uparrow\) 
& METEOR\,\(\uparrow\) 
& BLEU4\,\(\uparrow\) 
& SODA$\_$c \(\uparrow\) \\
\midrule
Vid2Seq       & \cmark & 47.10  & 9.30 & -- & 7.90 \\
DIBS          & \cmark & 44.40   & 7.51 & -- & 6.39 \\
\midrule
DDVC      & \xmark & 38.75  & 6.92 & 1.92 & 6.68 \\
\midrule
PDVC & \xmark & 29.69 & 5.56 & 1.40  & 4.92 \\
CM$^2$          & \xmark & 31.66  & 6.08 & 1.63 & 5.34 \\
MCCL          & \xmark & \underline{36.09}  & \textbf{6.53} & \underline{2.04} & 5.21 \\
E$^2$DVC       & \xmark & 34.26  & 6.11 & 1.68 & \underline{5.39} \\
\textbf{Ours} & \xmark & \textbf{39.18}  & \underline{6.14} & \textbf{2.10} & \textbf{7.06} \\
\bottomrule
\end{tabular*}
\end{table}

\begin{table}[t] 
\centering
\small
\setlength{\tabcolsep}{0pt}\renewcommand{\arraystretch}{1.1}
\caption{\textbf{ActivityNet Captions metrics.} \textbf{Bold} represents best and \underline{underline} represents second best performance. We do not directly compare to DDVC for fair comparison as they utilize GPT2 as their captioning head.}
\label{tab:anet_cap}
\begin{tabular*}{\columnwidth}{
    @{\hspace{4pt}}l@{\hspace{4pt}} 
    !{\vrule width 0.5pt} 
    @{\hspace{4pt}}c@{\hspace{4pt}} 
    !{\vrule width 0.5pt} 
    @{\extracolsep{\fill}} 
    c 
    c 
    c 
    @{\hspace{4pt}}c@{\hspace{4pt}}
}
\toprule
Method & \#PT
& CIDEr\,\(\uparrow\) 
& METEOR\,\(\uparrow\) 
& BLEU4\,\(\uparrow\) 
& SODA$\_$c\,\(\uparrow\) \\
\midrule
Vid2Seq       & \cmark & 30.10 & 8.50 & --   & 5.80 \\
DIBS          & \cmark & 31.89 & 8.93 & --   & 5.85 \\
\midrule
DDVC      & \xmark & 35.48 & 8.62 & 2.44 & 6.55 \\
\midrule
PDVC & \xmark & 29.97 & 8.06 & 2.21 & 5.92 \\
CM$^2$            & \xmark & 33.01 & 8.55 & 2.38 &\underline{6.18} \\
MCCL          & \xmark &\underline{34.92} &\textbf{9.05} &\textbf{2.68} & 6.16 \\
E$^2$DVC       & \xmark & 33.63 &\underline{8.57} &\underline{2.43} & 6.13 \\
\textbf{Ours} & \xmark &\textbf{35.04} & 8.45 & 2.36 &\textbf{6.45} \\
\bottomrule
\end{tabular*}
\end{table}

\subsection{Comparison with State-of-the-Art Methods}
\noindent\textbf{Captioning Performance.}
As shown in 
\Tref{tab:yc2_cap}
and 
\Tref{tab:anet_cap}, our method achieves strong captioning results on both datasets.
On YouCook2, our model surpasses all non-pretrained methods, achieving the best CIDEr and SODA$\_$c scores. Compared to MCCL which utilizes an external memory bank to enhance captioning diversity, our model is able to improve the CIDEr score by 3.09, BLEU4 by 0.06, and SODA$\_$c by 1.85. Even for METEOR, our method only lags behind by 0.39. 
In addition, compared to the latest non-memory bank method E$^2$DVC, our method surpasses it in CIDEr by 4.92.
While DDVC~\cite{liu-etal-2025-taskddvc_dvc6} employs a GPT-2 language model for caption generation, making it not directly comparable to ours, we still achieve comparable or better scores without utilizing an LLM based model, even surpassing it in CIDEr by 0.43. 
This demonstrates that our role-specific queries and Overlap Suppression Loss enable the caption head to produce coherent and semantically rich sentences purely from visual cues. 

On ActivityNet, our model performs favorably, and even achieving the best CIDEr and SODA$\_$c scores, surpassing MCCL in CIDEr by 0.12 and SODA$\_$c by 0.32.
These consistent gains confirm that our method contributes to precise temporal grounding and caption generation.

\noindent\textbf{Event Localization Performance.} We show localization results on YouCook2 and ActivityNet in~\Tref{tab:loc}. On YouCook2, our model surpasses all non-pretrained methods, achieving the best Recall, Precision, and F1 scores. Compared to the most recent non-memory bank method, E$^2$DVC, our method surpasses on Recall by 4.98, Precision by 0.51 and F1 by 3.39. On ActivityNet, our model's Recall surpasses E$^2$DVC by 0.68. We also find that, unlike previous methods where Recall is consistently lower than Precision, our model yields nearly matched Recall and Precision.
Since Recall is computed as predicted events over the number of ground-truth events, and Precision as predicted events over the number of model’s estimated events, this indicates that our event counter produces the number of events much closer to the ground truth.


\begin{table}[t]
\centering
\small
\setlength{\tabcolsep}{0pt}\renewcommand{\arraystretch}{1.1}
\caption{\textbf{Localization metrics on YouCook2 and ActivityNet.} \textbf{Bold} represents best and \underline{underline} represents second best performance. We do not directly compare to DDVC for fair comparison as they utilize GPT2 as their captioning head.}
\label{tab:loc}
\begin{tabular*}{\columnwidth}{
    @{\hspace{4pt}}c@{\hspace{4pt}}
    !{\vrule width 0.5pt}
    @{\hspace{4pt}}c@{\hspace{4pt}} 
    !{\vrule width 0.5pt}
    @{\extracolsep{\fill}} 
    c 
    c 
    c @{\hspace{8pt}} 
    !{\vrule width 0.5pt}
    @{\hspace{8pt}} c 
    c 
    c 
    @{\hspace{4pt}} 
}
\toprule
& & \multicolumn{3}{c}{YouCook2} & \multicolumn{3}{c}{ActivityNet} \\
Method & \#PT
& Rec.\,$\uparrow$ & Pre.\,$\uparrow$ & F1\,$\uparrow$ 
& Rec.\,$\uparrow$ & Pre.\,$\uparrow$ & F1\,$\uparrow$ \\
\midrule
Vid2Seq       & \cmark & 27.90 & 27.80 & 27.84 & 52.70 & 53.90 & 53.29 \\
DIBS          & \cmark & 26.24 & 39.18 & 31.43 & 53.14 & 58.31 & 55.61 \\
\midrule
DDVC     & \xmark & 30.81 & 37.25 & 33.73 & 54.77 & 57.54 & 56.12 \\
\midrule
PDVC & \xmark & 22.89 & 32.37 & 26.81 & 53.27 & 56.38 & 54.78 \\
CM$^2$           & \xmark & \underline{24.76} & 33.38 & 28.43 & 53.71 & 56.81 & 55.21 \\
MCCL          & \xmark & --    & --    & --    & 53.19 & \underline{57.36} & 55.23 \\
E$^2$DVC       & \xmark & 24.36 & \underline{34.75} & \underline{28.64} & \underline{54.67} & \textbf{57.70} & \textbf{56.14} \\
\textbf{Ours} & \xmark & \textbf{29.34} & \textbf{35.26} & \textbf{32.03} & \textbf{55.35} & 55.65 & \underline{55.50} \\
\bottomrule
\end{tabular*}
\end{table}

\subsection{Ablation Experiments}

We conduct ablation studies to evaluate the impact of each component and design choice in our framework. 
All experiments are performed on the YouCook2 dataset with PDVC~\cite{Wang2021pdvc_dvc2} set as our baselines unless otherwise specified.

\begin{table}[t]
\centering
\footnotesize
\setlength{\tabcolsep}{3pt}
\renewcommand{\arraystretch}{1.05}
\caption{~\textbf{The ablation results of different components.} 
RSQI denotes \textit{Role Specific Query Initialization}, CTCA denotes \textit{Cross-Task Contrastive Loss}, 
OSL denotes \textit{Overlap Suppression Loss}, and CG denotes \textit{Concept Guider}. 
Note that CTCA cannot be applied when RSQI is not used.
}
\label{tab:ablation_module}
\begin{tabularx}{\columnwidth}{@{}
    Y      
    Y       
    Y
    Y
    S[table-format=2.2] 
    S[table-format=1.2] 
    S[table-format=1.2]
    S[table-format=1.2] 
    S[table-format=2.2] 
@{}
}
\toprule
RSQI & CTCA & OSL & CG &
 \multicolumn{1}{c}{C$\uparrow$} &
 \multicolumn{1}{c}{M$\uparrow$} &
 \multicolumn{1}{c}{B@4$\uparrow$} &
 \multicolumn{1}{c}{S$\uparrow$} &
 \multicolumn{1}{c}{F1$\uparrow$} \\
\midrule
   \xmark & \xmark  & \xmark & \xmark & 29.69 & 5.56 & 1.40 & 5.39 & 26.81 \\
   \cmark & \xmark  & \xmark & \xmark & 32.33 & 5.66 & 1.66 & 5.43 & 27.00 \\
   \xmark & \xmark  & \cmark & \xmark & 33.60 & 5.41 & 1.46 & 6.79 & 31.22 \\
   \xmark & \xmark  & \xmark & \cmark & 31.40 & 5.58 & 1.58 & 5.62 & 27.69 \\
   \cmark & \cmark  & \xmark & \xmark & 34.48 & 5.82 & 1.88 & 5.58 & 27.59 \\
   \cmark & \xmark  & \cmark & \xmark & 32.72 & 5.60 & 1.49 & 6.73 & 31.98 \\
   \cmark & \xmark  & \xmark & \cmark & 33.36 & 5.63 & 1.63 & 6.03 & 28.94 \\
   \xmark & \xmark  & \cmark & \cmark & 33.11 & 5.32 & 1.58 & 6.91 & 31.63 \\
   \cmark & \xmark  & \cmark & \cmark & 34.92 & 5.66 & 1.62 & 6.83 & 30.58 \\
   \cmark & \cmark  & \cmark & \xmark & 36.63 & 5.80 & 1.87 & 6.95 & 30.62 \\
   \cmark & \cmark  & \xmark & \cmark & 36.93 & 5.73 & 2.02 & 6.05 & 28.59 \\
    \rowcolor{gray!20}\cmark &\cmark   &\cmark  &\cmark  & \best{39.18} & \best{6.14} & \best{2.10} & \best{7.06} & \best{32.03} \\
\bottomrule
\end{tabularx}
\end{table}





\begin{table}[t]
\centering
\small
\setlength{\tabcolsep}{0pt} \renewcommand{\arraystretch}{1.1}
\caption{\textbf{Ablation results of varying $\gamma$ on OSL.} $\gamma$ is a hyperparameter that adjusts the influence of IoU with respect to the ground truth.}
\label{tab:gamma_abl}
\begin{tabularx}{\columnwidth}{
    Y
    !{\vrule width 0.5pt} 
    Y
    Y
    Y 
    Y 
    Y
    @{\hspace{4pt}}Y@{\hspace{4pt}} %
} 
\toprule
 $\gamma$
& CIDEr\,\(\uparrow\) 
& SODA$\_$c \(\uparrow\) 
& Rec.\,\(\uparrow\) 
& Pre.\,\(\uparrow\) 
& F1\,\(\uparrow\) \\
\midrule
0.15 & 37.00 & 6.69 & 29.65 & \textbf{35.44} & 32.28 \\
\rowcolor{gray!20} \textbf{0.25} & \textbf{39.18} & 7.06 & 29.34 & 35.26 & 32.03 \\
0.35 & 37.49 & \textbf{7.16} & \textbf{30.80} & 35.28 & \textbf{32.88} \\
0.45 & 35.76 & 6.94 & 30.40 & 34.43  & 32.28 \\
\bottomrule
\end{tabularx}
\end{table}

\begin{table}[t]
\centering
\small
\setlength{\tabcolsep}{0pt} \renewcommand{\arraystretch}{1.1}
\caption{\textbf{Ablation results with and without $\alpha$.} $\alpha$ controls whether the IoU is compared with the ground truth in OSL.}
\label{tab:alpha_abl}
\begin{tabularx}{\columnwidth}{
    Y     
    !{\vrule width 0.5pt} 
    Y
    Y
    Y
    Y
    Y
} 
\toprule
 $\alpha$
& CIDEr\,\(\uparrow\) 
& SODA$\_$c \(\uparrow\) 
& Rec.\,\(\uparrow\) 
& Pre.\,\(\uparrow\) 
& F1\,\(\uparrow\) \\
\midrule
 W/o  $\alpha$  & 34.81 & \textbf{7.20} & \textbf{31.01} & 33.82  & \textbf{32.35} \\
 \rowcolor{gray!20} \textbf{W/ $\alpha$}  & \textbf{39.18} & 7.06 & 29.34 & \textbf{35.26} & 32.03 \\
\bottomrule
\end{tabularx}
\end{table}


\begin{figure*}[t]
    \centering
    \includegraphics[width=1\linewidth]{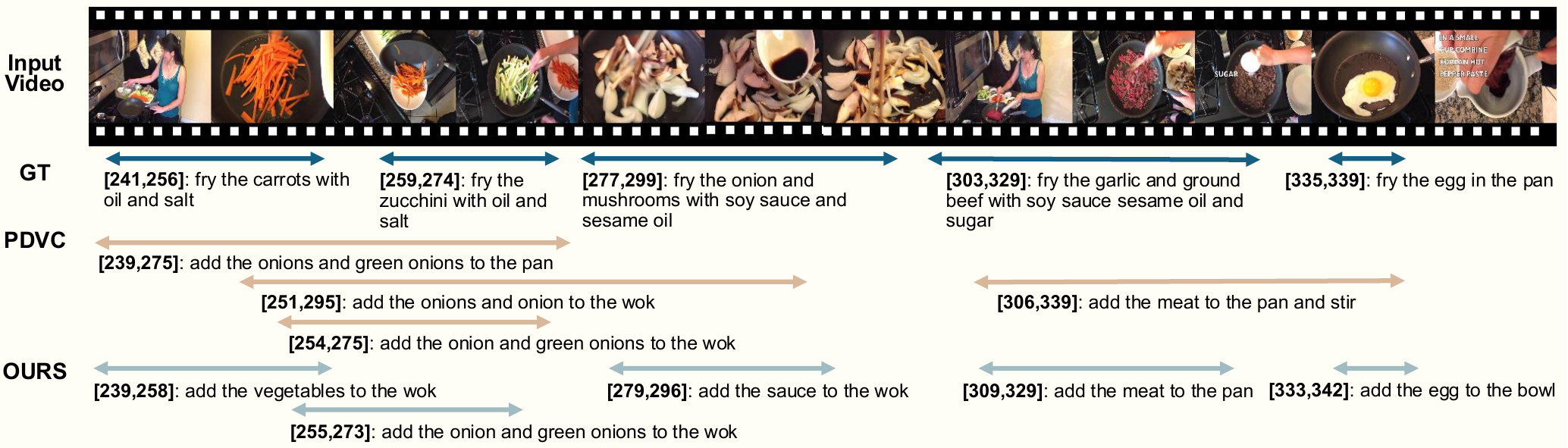}
    \caption{\textbf{Qualitative results of dense video captioning on YouCook2.} We compare the localization and captioning results with the ground truth, the baseline (PDVC), and ours. Each arrow represents localization boundaries and its corresponding caption is provided below. We find that while PDVC captures redundant events with the same captions, our model avoids this phenomenon. }
    \label{fig:qual}
\end{figure*}

\noindent\textbf{Effect of Each Component.}
As shown in~\Tref{tab:ablation_module}, we ablate each component of our model, Role Specific Query Initialization (RSQI), Cross-Task Contrastive Alignment (CTCA), Overlap Suppression Loss (OSL), and Concept Guider (CG).
Using only RSQI allows each query to better specialize for its corresponding task, which in turn leads to noticeable improvements in both CIDEr and F1.
When only OSL is applied, the provided query guidance significantly boosts the F1 score. 
Adding CG alone enhances the expressiveness of the generated captions, thereby improving the CIDEr score. 
Finally, 
our model with all components achieves the best performance across all metrics.

\noindent\textbf{Effect of Overlap Suppression Loss.}
To ablate the effect of OSL, we set $\gamma$ from 0.15 to 0.45. $\gamma$ is a hyperparameter that adjusts the influence of IoU with the ground truth.
As shown in~\Tref{tab:gamma_abl}, setting $\gamma$ to 0.25, which is the middle of the range, we achieve a good balance among different metrics, yielding the highest CIDEr score of 39.16 and competitive results in other metrics. This suggests that $\gamma=0.25$ effectively manages the trade-off between Recall and Precision.
~\Tref{tab:alpha_abl} shows the performance of OSL with and without $\alpha$. Without $\alpha$, we just penalize the IoU between two distinct predicted boundaries. The model achieves higher Recall without the $\alpha$ term, whereas adding the $\alpha$ term improves Precision and CIDEr.
Since DVC should achieve strong performance in both captioning and localization to ensure accurate and coherent video understanding, OSL with $\alpha$ at $\gamma=0.25$ demonstrates an effective balance, yielding strong results across both metrics.


\begin{table}[t]
\centering
\small
\setlength{\tabcolsep}{0pt} \renewcommand{\arraystretch}{1.1}
\caption{
~\textbf{Ablation results varying numbers of queries.} 
$\mathrm{N}_{q}$ denotes the number of queries.
}
\label{tab:ablation_per_query}
\begin{tabularx}{\columnwidth}{
    >{\hspace{5pt}}c<{\hspace{5pt}}
    !{\vrule width 0.5pt}  
    Y Y Y Y Y
} 
\toprule
$\mathrm{N}_{q}$
& CIDEr$\uparrow$
& METEOR$\uparrow$
& BLEU4$\uparrow$
& SODA$_c$$\uparrow$
& F1$\uparrow$ \\
\midrule
10        & 28.72 & 4.78 & 1.17 & 6.79 & 29.64 \\
20        & 35.21 & 5.69  & 1.60 & 7.16 & 32.55 \\
30        & 34.29 & 5.65  & 1.56 & 7.03 & 31.96 \\
40        & 34.59 & 5.77  & 1.58 & \textbf{7.19} & \textbf{33.56}  \\
\rowcolor{gray!20} \textbf{50} & \textbf{39.18} & \textbf{6.14}  & \textbf{2.10} & 7.06 & 32.02 \\
60        & 36.71 & 5.90  & 1.77 & 6.53 & 30.64 \\
70        & 36.24 & 5.88  & 1.83 & 6.81 & 32.01 \\
80        & 34.44 & 5.75  & 1.65 & 6.66 & 31.32 \\
90        & 36.83 & 5.96 & 1.94 & 6.28 & 29.34 \\
100       & 34.82 & 5.66  & 1.79 & 6.47 & 27.83 \\
\bottomrule
\end{tabularx}
\end{table}



\noindent\textbf{Query number.}
We ablate the number of queries to identify the optimal query number for our method in ~\Tref{tab:ablation_per_query}.
PDVC sets the number of queries to 100 for YC2, while ours uses $2 \times N_q$ queries due to the two query sets.
When the number of queries is small, the model tends to miss events, resulting in lower Recall and limited caption coverage.
As the number increases, performance improves and peaks at around 50 queries, where captioning quality and event localization are well balanced.
Using too many queries leads to redundant or overlapping proposals that slightly reduce temporal precision.
We find $N_q=50$ to be optimal across all metrics.



\subsection{Qualitative Results}

We show qualitative results in~\Fref{fig:qual}.
The baseline PDVC captures redundant overlaps that generate the same captions.
In contrast, our method effectively suppresses unnecessary overlapping events, demonstrating that the Role Specific Query and OSL help prevent both repetitive captions and redundant overlaps, ultimately enhancing localization and captioning quality. 

\section{Conclusion}
\label{sec:conclusion}

In this work, we tackle Dense Video Captioning through a query-based framework that focuses on improving how learnable queries are trained for localization and captioning. 
With our model ROS-DVC, we introduce three key components, Role Specific Query, Overlap Suppression Loss, and Concept Guider, to enable task-aware query optimization, suppress redundant detections, and enhance concept-level reasoning. Experimental results on YouCook2 and ActivityNet Captions demonstrate that our approach effectively mitigates task interference and redundancy, achieving more precise localization and semantically coherent captions. Overall, ROS-DVC provides a simple yet effective solution for query-based DVC.


{
    \small
    \bibliographystyle{ieeenat_fullname}
    \bibliography{main}
}


\end{document}